\documentclass[10pt,twocolumn,letterpaper]{article}
\usepackage[accsupp]{axessibility}
\usepackage{wacv}              
\usepackage{graphicx}
\usepackage{amsmath}
\usepackage{amssymb}
\usepackage{booktabs}

\usepackage{multirow}
\usepackage[pagebackref,breaklinks,colorlinks]{hyperref}
\def\bx{\mathbf{x}}
\def\tbx{\widetilde{\mathbf{x}}}

\usepackage{algorithm,algpseudocode,mathrsfs,subcaption}
\usepackage[capitalize]{cleveref}
\crefname{section}{Sec.}{Secs.}
\Crefname{section}{Section}{Sections}
\Crefname{table}{Table}{Tables}
\crefname{table}{Tab.}{Tabs.}

\def\wacvPaperID{1020} 
\def\confName{WACV}
\def\confYear{2025}

\begin{document}

\title{Decomposed Distribution Matching in Dataset Condensation}

\author{Sahar Rahimi Malakshan, Mohammad Saeed Ebrahimi Saadabadi,\\
Ali Dabouei, and Nasser M. Nasrabadi\\
{\tt\small {sr00033, me00018, Ad0046}@mix.wvu.edu, nasser.nasrabadi@mail.wvu.edu}
}

\maketitle

\begin{abstract}
   Dataset Condensation (DC) aims to reduce deep neural networks training efforts by synthesizing a small dataset such that it will be as effective as the original large dataset. Conventionally, DC relies on a costly bi-level optimization which prohibits its practicality. Recent research formulates DC as a distribution matching problem which circumvents the costly bi-level optimization. However, this efficiency sacrifices the DC performance.
   To investigate this performance degradation, we decomposed the dataset distribution into content and style. Our observations indicate two major shortcomings of: 1) style discrepancy between original and condensed data, and 2) limited intra-class diversity of condensed dataset.
   We present a simple yet effective method to match the style information between original and condensed data, employing statistical moments of feature maps as well-established style indicators.
   Moreover, we enhance the intra-class diversity by maximizing the Kullback–Leibler divergence within each synthetic class, \ie, content.
   We demonstrate the efficacy of our method through experiments on diverse datasets of varying size and resolution, achieving improvements of up to 4.1\% on CIFAR10, 4.2\% on CIFAR100, 4.3\% on TinyImageNet, 2.0\% on ImageNet-1K, 3.3\% on ImageWoof, 2.5\% on ImageNette, and 5.5\% in continual learning accuracy. 
   \href{https://github.com/SaharR1372/DM_Style_matching}{Code}

\end{abstract}

\vspace{-5pt}
\section{Introduction}
\label{sec:intro}
\vspace{-5pt}
In response to the challenges imposed by the sheer amount of data in large-scale datasets, \eg, storage and computational burden, the concept of Dataset Condensation (DC) was introduced \cite{wang2018dataset,feng2023embarrassingly,zhao2023dataset,wang2022cafe,cazenavette2022dataset}.
Pioneered by Wang \etal \cite{wang2018dataset}, DC utilizes a nested optimization to synthesize a small dataset that retains the effectiveness of the original dataset. Despite inspiring, their proposal was computationally intensive and infeasible for large-scale setups \cite{feng2023embarrassingly,zhao2023dataset,wang2022cafe,cazenavette2022dataset}. Therefore, follow-up studies \cite{cazenavette2022dataset,zhao2021DC, lee2022dataset, wang2022cafe, zhao2021DSA} try to circumvent the nested optimization of \cite{wang2018dataset} by matching the training trajectories \cite{cazenavette2022dataset,du2023minimizing} or gradients \cite{zhao2021DC, lee2022dataset, wang2022cafe, zhao2021DSA} of surrogate models trained on condensed and original datasets. Although promising, relying on computationally extensive bi-level optimization, \ie, an inner optimization for model updates and an outer one for condensed data updates, limits their practicality \cite{cazenavette2022dataset, zhao2021DC, zhao2023dataset, lee2022dataset}.
\begin{figure}
  \centering
   \includegraphics[width=1.0\linewidth]{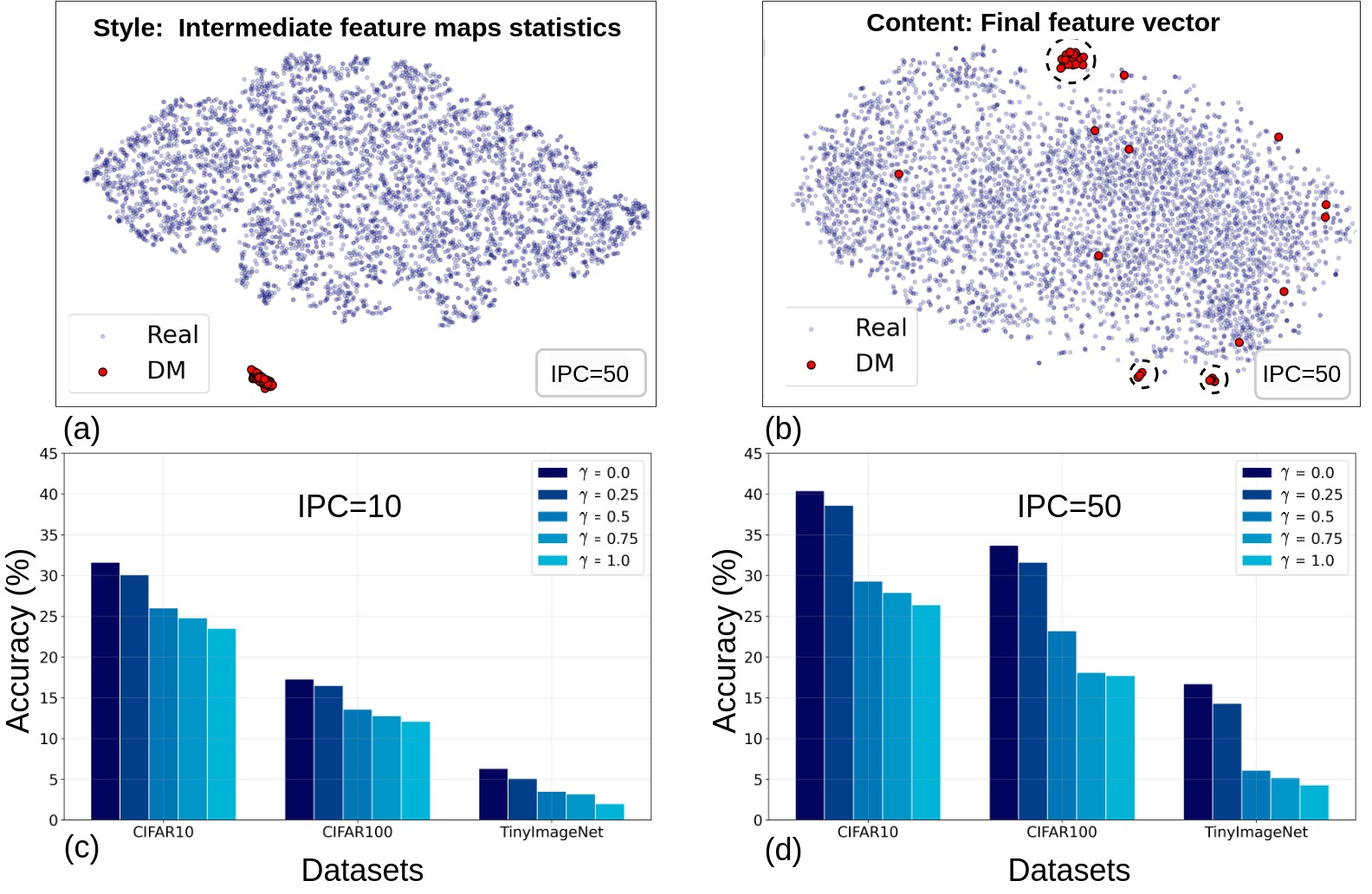}
   \caption{(a, b) 2D t-SNE visualizations of original and condensed images learned by DM \cite{zhao2023dataset} for CIFAR10 with IPC=50 in randomly chosen category. (a) Style statistics (concatenation of mean and variance) from the first layer's feature map, highlighting a significant style discrepancy. (b) Final features of the DNN, showing limited diversity of instances learned by DM. (c, d) Illustrating the negative effect of style discrepancy on performance. During training the style of samples from Herding is \cite{rebuffi2017icarl} drifted toward that of DM \cite{zhao2023dataset}, with $\gamma$ representing the drift ratio.}
   \label{fig:styleStat}
   \vspace{-5pt}
\end{figure}

Recently, Zhao \etal \cite{zhao2023dataset} circumvent the bi-level optimization by leveraging the distance-preserving property of representations obtained from randomly sampled Deep Neural Networks (DNNs), \ie, DNN with random weights \cite{saxe2011random}. Specifically, they formulate DC as a distribution matching problem between the original and condensed datasets in the embedding of randomly sampled DNNs, dubbed DM \cite{zhao2023dataset}.
Utilizing random DNNs bypasses the inner optimization of bi-level methods \cite{zhao2023dataset}. Consequently, DM severely reduces the computational cost/time of DC since it only updates the condensed data.
For instance, for 50 Images Per Class (IPC), DM condenses CIFAR10 $45 \times$ faster than bi-level optimization methods like \cite{ zhao2021DSA,zhao2021DC}.

Despite the efficiency of DM \cite{zhao2023dataset}, its performance lags behind that of bi-level methods \cite{zhao2021DC, lee2022dataset, wang2022cafe, zhao2021DSA, cazenavette2022dataset}. 
To study this deficiency, motivated by the literature on distribution matching \cite{zhang2022exact,zhou2023semi,kang2022style,zhang2019few,li2016revisiting,kundu2020universal,li2019learning,huang2017arbitrary,li2019optimal},
we decompose the dataset distribution into two major factors: 1) style, including attributes like texture, and color, and 2) content, which encompasses the semantic information \cite{huang2017arbitrary,li2019learning,gatys2015neural,gatys2015texture}. 
We trained a Convolutional Neural Network (CNN) using original and condensed data learned by DM \cite{zhou2021domain}. 
Then, we explored the content and style discrepancy between original and condensed data in the embedding of the trained CNN, shown in Figure \ref{fig:styleStat}. 

The first and second moments of the intermediate feature maps of CNNs, \ie, mean and variance, capture the style of the input image \cite{li2017demystifying,kalischek2021light,zhang2022exact}.
As Figure \ref{fig:styleStat}a illustrates, there is a significant style discrepancy across original and condensed data. Specifically, despite the similar content (same category), original and condensed data represent distinct styles. 
Style gap between training and testing results in severe performance degradation due to the DNNs bias toward style \cite{nam2021reducing,hermann2020origins,kang2022style,zhou2021domain,zhao2022style}.
Furthermore, Figure \ref{fig:styleStat}c, and d illustrate the effect of this style gap in the performance obtained from condensed datasets.
Specifically, during training, the style of samples from Herding coreset selection \cite{rebuffi2017icarl} \ie, original styles, is drifted towards that of DM \cite{zhao2023dataset}.
Please refer to Section A of Supplementary Materials for more details. As style drifts from the real, the performance decreases, reflecting the importance of style alignment between condensed and real datasets; in line with previous studies on dataset distribution \cite{geirhos2018imagenet,baker2018deep,zhou2021domain,zhang2022exact,nam2021reducing}.

Content information of the input image is reflected in the final feature embedding of DNNs \cite{huang2017arbitrary,saito2018maximum,ma2015hierarchical,gretton2012kernel}. 
Figure \ref{fig:styleStat}b compares the t-SNE visualization of feature vectors of original and condensed data, showing no evident content gap between them.
However, it reveals a lower intra-class diversity of condensed samples than the original. Specifically, condensed instances form local clusters in the embedding space, reflecting similar information, \ie, low intra-class diversity \cite{trabucco2023effective}.
DM's training objective \cite{zhao2023dataset}
explicitly promotes the content alignment between real and condensed datasets \cite{sajedi2023datadam,zhao2023dataset} but discards the diversity.
Thus, condensed data fails to adequately represent the original dataset's extensive variability, leading to overfitting when used as a source of training data \cite{zhang2023dataset,aljundi2019gradient}.

Prior works for improving DM \cite{sajedi2023datadam,zhao2023improved,zhang2024m3d} either incur significant computational costs to its framework or employ a restricted spatial supervision that reduces the generalization \cite{deng2024exploiting,khaki2024atom}. 
In this study, our key insight is that condensed data should: 1) express the distribution of original data in both style and content, 2) consist of diverse informative samples, and 3) be synthesized without a bi-level learning regime to be applicable to large scale setups.
Concerning style disparity, in addition to the content alignment of DM, we propose a Style Matching (SM) module. SM module leverages well-established style indicators of feature map moments and correlations to align the style across original and condensed data. Our proposal leverages feature maps from a randomly sampled DNNs, \ie  adhering to the computationally efficient framework of DM, to match the style between real and condensed sets. 

To encourage intra-class diversity, we employ a criterion based on Kullback–Leibler (KL) divergence \cite{kullback1951information} to penalize samples that form a local cluster.
Our proposal works in the embedding of a random DNN and encourages intra-class diversity while maintaining the plausibility of the samples and the computational efficiency of the DM framework.
Our method demonstrates significant improvements across diverse datasets with low, medium, and high resolutions, including CIFAR10, CIFAR100, Tiny ImageNet, and ImageNet-1K, affirming its scalability and generalization from small to large scale datasets.
Also, we show the generalization of the proposed method by evaluating on both simple ConvNet and the more sophisticated ResNet architectures. 
The contributions of the paper are as follows:
\begin{itemize}
    \item We decompose the distribution matching framework in DC into two major factors: content and style, and reveal the shortcomings of DC in these factors.
    \item We identify the issue of the style gap between original and condensed data. Then we propose an optimization based on matching statistical moments
    of feature maps to reduce this style disparity.
    \item We identify the issue of limited intra-class diversity of the distribution matching process in DC. Then, we propose an optimization method specifically tailored to increase the intra-class diversity by penalizing the synthesized samples that express similar information.
\end{itemize}
\vspace{-8pt}
\section{Related Work}
\subsection{Coreset Selection}
Coreset or instance selection is a heuristic method that approximates the full dataset by a small subset \cite{feldman2020introduction}.
For instance, random selection \cite{rebuffi2017icarl} chooses samples arbitrarily; Herding \cite{castro2018end, belouadah2020scail} selects samples nearest to each class's cluster center;
and Forgetting \cite{toneva2019empirical} identifies samples that are easily forgotten during training.
Despite the advances in coreset selection methods, they fail to scale into large-scale setups due to the computational deficiency \cite{yu2023dataset,guo2022deepcore}. Moreover, the heuristic criteria cannot guarantee the optimal solution for down-stream tasks \cite{zhao2023improved}.
Dataset condensation offer an alternative approach by synthesizing condensed data that can overcome the limitations of coreset selection methods \cite{yu2023dataset}.

\subsection{Dataset Condensation }
Dataset Condensation (DC) or dataset distillation synthesizes condensed datasets that retain the learning properties of larger originals, enabling efficient model training with reduced data \cite{wang2018dataset}. This technique has applications in continual learning \cite{sangermano2022sample,rosasco2021distilled}, privacy protection \cite{carlini2022no,dong2022privacy}, and neural architecture search \cite{cui2022dc}, among others.
Wang \etal \cite{wang2018dataset} introduced DC, framing it as a meta-learning problem where network parameters are optimized as a function of synthetic data to minimize the training loss on real datasets.
Building on this foundation, subsequent studies have leveraged surrogate objectives to address the unrolled optimization challenges inherent in meta-learning framework. Notably, gradient matching methods \cite{zhao2021DC,lee2022dataset,zhao2021DSA,du2024sequential,kim2022dataset} align DNN gradients between original and condensed datasets, while trajectory matching approaches \cite{cazenavette2022dataset,du2023minimizing,cui2023scaling} align the DNNs' parameter trajectories.
Although promising, their reliance on computationally intensive bi-level optimization hinders their applicability to large-scale setups \cite{zhang2023accelerating,feng2023embarrassingly}.

To address these limitations, Zhao \etal \cite{zhao2023dataset} formulate DC as a distribution matching problem between the original and condensed datasets within the embeddings of randomly sampled DNNs. Specifically, DM \cite{zhao2023dataset} aligns the feature distributions of condensed and original datasets by matching their penultimate layer feature representations.
However, DM \cite{zhao2023dataset} sacrifices the performance to maintain the efficiency.
Thus, strategies such as IDM by Zhao \etal \cite{zhao2023improved} and Datadam by Sajedi \etal \cite{sajedi2023datadam} have been introduced to enhance DM.
IDM \cite{zhao2023improved} employs semi-trained models, and class-aware regularization, to improve DM performance. However, it diverge from efficient optimization based on randomly sampled DNNs.
DataDAM \cite{sajedi2023datadam}, improves DM by using spatial supervision to align the attention maps between real and synthetic datasets. However, such restricted spatial supervision leads to the generalization reduction \cite{deng2024exploiting,khaki2024atom}. Also , CAFE \cite{wang2022cafe} aligns feature distributions of condensed and real datasets across multiple DNN layers using a dynamic bi-level optimization framework. However, it diverges from DM efficient framework, leading to considerable computation cost.

\begin{figure*}
    \centering
    
    \includegraphics[width=0.9\textwidth]{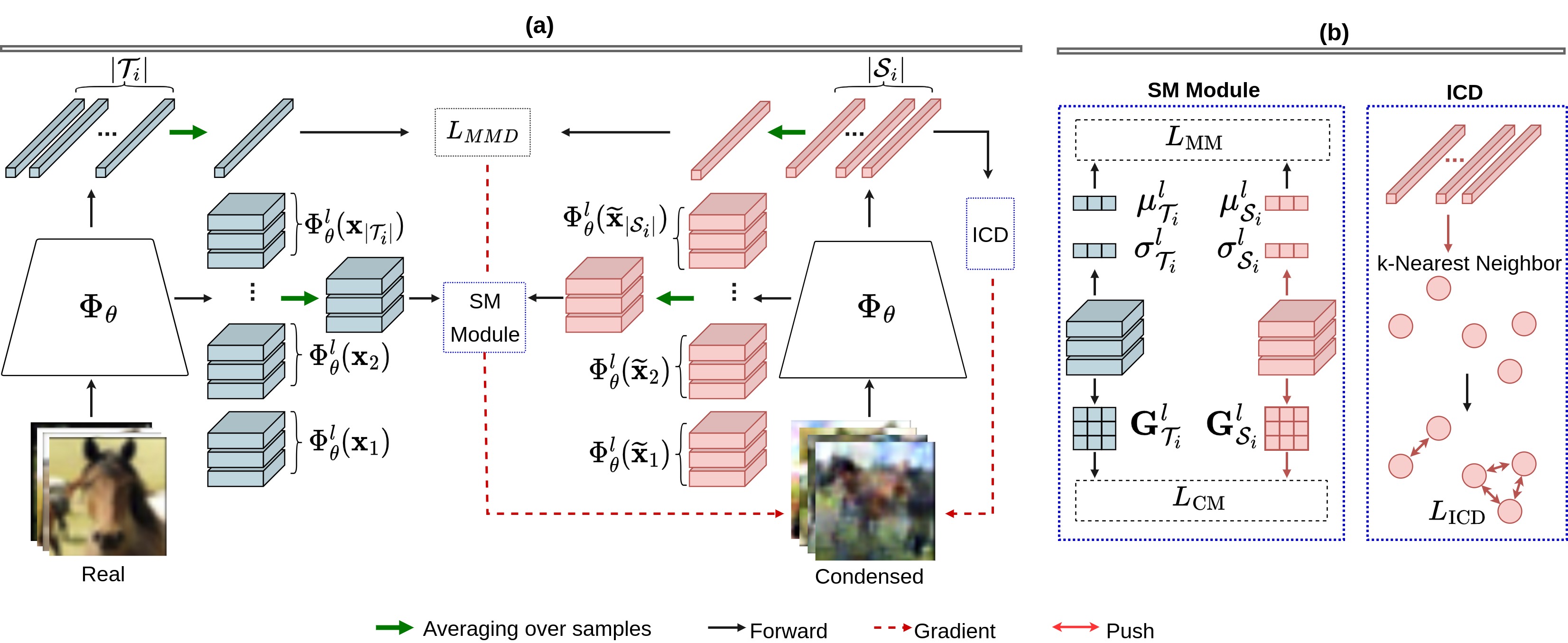}

    \caption{ (a) Visualization of the proposed method, which includes a Style Matching (SM) module and Intra-Class Diversity (ICD) components. (b) SM module includes Moments Matching (MM) and Correlation Matching (CM) losses to reduce style discrepancies between real and condensed sets by using the \ie, mean and variance of feature maps as well as correlation among feature maps captured by the Gram matrix in a DNN across different layers. Meanwhile, the ICD component enhances diversity within condensed sets by pushing each condensed sample away from its $k$ nearest intra-class neighbors.}

    \label{fig:overview}
    \vspace{-13pt}
\end{figure*}

\subsection{Style}

Style of an image encompasses its visual attributes such as texture and color \cite{gatys2015neural}, which are widely represented by the characteristics of intermediate feature maps \cite{huang2017arbitrary,li2016revisiting,li2017demystifying,ulyanov2016instance,zhou2021domain}.
Assuming a Gaussian prior for features, the first and second moments, \ie, the mean and variance, of DNN feature maps are well-established style indicators \cite{huang2017arbitrary,li2016revisiting,li2017demystifying,ulyanov2016instance,zhou2021domain}.
Furthermore,
second order moment between feature activations, captured by the Gram matrix is another widely used style indicator \cite{gatys2016image, gatys2015texture,li2017universal}. DNNs show strong bias toward input style, leading to severe performance degradation when the style of the training data is not align with that of test \cite{nam2021reducing,hermann2020origins,kang2022style,zhou2021domain,zhao2022style}. Style features have been widely used in style transfer \cite{gatys2016image, gatys2015texture,li2017universal}, domain adaptation and generalization \cite{zhou2021domain,ghifary2016scatter,zhou2024mixstyle}, amongst others, showcasing the importance of training and testing style alignment. However, previous DM-based studies have been overlook the signficant role of style, leading to style gap between condensed and original data, as shown in Figure \ref{fig:styleStat}a. In this work, we aim to reduce this style gap by employing well-established style indicators in DM-framework.

\section{Proposed Method}
\subsection{Notation}
In this paper, we use lowercase letters (\eg., $x$) to denote scalars, lowercase boldface (\eg, $\mathbf{x}$) to denote vectors, uppercase letters (\eg, $X$) to denote functions, uppercase boldface (\eg, $\mathbf{X}$) to denote matrices and uppercase calligraphic symbols (\eg, $\mathcal{X}$) to denote sets.

\subsection{Preliminary} \label{sec:preDM}
DC aims to learn a small condensed dataset $\mathcal{S}=\{(\widetilde{\mathbf{x}}_1,y_1),\dots,(\widetilde{\mathbf{x}}_{|\mathcal{S}|},y_{|\mathcal{S}|})\}$ from a large-scale real dataset $\mathcal{T}=\{(\mathbf{x}_1,y_1),\dots,(\mathbf{x}_{|\mathcal{T}|},y_{|\mathcal{T}|})\}$, such that $|\mathcal{S}|\ll |\mathcal{T}|$ and $\mathcal{S}$ preserve essential information presented in $\mathcal{T}$ \cite{wang2018dataset}. Specifically, DC seeks to learn $\mathcal{S}$ in a way that an arbitrary learning function trained on $\mathcal{S}$ can have similar performance as that trained on $\mathcal{T}$ \cite{wang2018dataset,zhao2023dataset}:
\begin{equation}\label{eq:combined}
\resizebox{0.89\linewidth}{!}{$
\begin{aligned}
    \mathcal{S}^{*} & = \underset{\mathcal{S}}{\arg\min } \:\:  \mathbb{E}_{(\bx,y)\sim P_D} \left[ \left|{L}\left(\Phi_{\theta_{\mathcal{T}}}(\bx),  y\right)-{L}\left(\Phi_{\theta_{\mathcal{S}}}(\bx), y\right)\right|\right],
\end{aligned}$}
\end{equation}
where $\bx$ is a sample from real image distribution $P_D$, and $y$ is its corresponding label. $\Phi_{\theta}: \mathbb{R}^{q}\rightarrow \mathbb{R}^{d}$ denotes a mapping, \ie, DNN, with trainable parameter $\theta$, that maps $\bx \in \mathbb{R}^q$ to $d$-dimensional embedding space. For RGB input $q$ is $3\times h \times w$.
Furthermore, $\theta_{\mathcal{T}}$, and $\theta_{\mathcal{S}}$ are two samples from the distribution of $\theta$ trained on $\mathcal{T}$, and $\mathcal{S}$, respectively. Finally, $L$ represents the learning objective function, \eg, empirical risk. 

Intuitive approach to solving the optimization in Equation \ref{eq:combined} is to use a bi-level learning regime by optimizing $\mathcal{S}$ and $\theta$ in turn \cite{zhao2021DC,cazenavette2022dataset, lee2022dataset,wang2022cafe,zhao2021DSA}.
However,
the nested loop of alternately optimizing $\theta$ and $\mathcal{S}$ is computationally intensive and scales poorly to large datasets and complex architectures \cite{cazenavette2022dataset, zhao2021DC, lee2022dataset,feng2023embarrassingly,zhao2023dataset}.
Inspired by the observation of Giryes \etal \cite{giryes2016deep} that a $\Phi_{\theta}$ with random $\theta$ performs a distance-preserving embedding, Zhao \etal \cite{zhao2023dataset} demonstrate that the validity of $\mathcal{S}$ can also be guaranteed even with random $\theta$.
Specifically, \cite{zhao2023dataset} reformulates the DC objective as a distribution matching problem in the embedding of random DNNs. 
\cite{zhao2023dataset} enforces the alignment between feature distribution of $\mathcal{S}$ with that of $\mathcal{T}$ in the embedding of random mappings $\Phi_\theta$:
\begin{equation}\label{lossGeneral}
\resizebox{0.55\linewidth}{!}{$
\begin{aligned}
    \mathcal{S}^{*}= \underset{\mathcal{S}}{\arg\min } \: \:\mathbb{E}_{\theta\sim\Theta}[D(\mathcal{S},\mathcal{T};\Phi_\theta)],
\end{aligned}$}
\end{equation}
where ${\Theta}$ is the distribution of $\theta$, \ie, the distribution used to initialize network parameters, and $D$ is an arbitrary metric measuring the divergence between the two distributions. 

Equation \ref{lossGeneral} circumvents the nested loop of the bi-level optimization methods by solely updating $\mathcal{S}$, significantly reducing DC computational cost. 
This method is known as Distribution Matching (DM) for dataset condensation.
Taking Maximum Mean Discrepancy (MMD) \cite{gretton2012kernel} as $D$, the objective function in DM \cite{zhao2023dataset} is defined as:
\begin{equation}\label{originalDM}
\resizebox{0.98\linewidth}{!}{$
\begin{aligned}
    L_{{MMD}}
= &  \: \mathbb{E}_{\theta\sim\Theta}\bigl[\textstyle\sum_{i=0}^{c-1}\bigl\| \frac{1}{|\mathcal{S}_i|}\sum_{\tbx \in \mathcal{S}_i}\Phi_{\theta}(\tbx) - \frac{1}{|\mathcal{T}_i|}\sum_{\bx \in \mathcal{T}_i}\Phi_{\theta}(\mathbf{x}) \bigr\|^2\bigr],
\end{aligned}$}
\end{equation}
where $\mathcal{S}_i$ and $\mathcal{T}_i$ are the subsets of condensed and real datasets, respectively, for the $i$-th class, and $c$ is the number of classes.

Despite the efficiency of DM,
this expedited learning comes at the cost of reduced performance, \eg, an 8\% performance reduction in CIFAR100 ($|\mathcal{T}|=50000$) compared to DSA \cite{zhao2021DSA} (a bi-level optimization method) when condensing all images in a class into 10 instances ($|\mathcal{S}|=1000$).  
To study this performance degradation, we decompose the feature distribution into its major factors, \ie style and content components \cite{huang2017arbitrary,li2019learning,multidomain3,zhou2021domain,zhou2023semi,gatys2015neural,gatys2015texture}. Style expresses attributes such as texture, color, and smoothness, while the content captures semantic information.
Our exploratory experiments, depicted in Figure \ref{fig:styleStat}, reveal two major shortcomings in the synthetic dataset $\mathcal{S}$ learned by DM compared to $\mathcal{T}$: (1) considerable style discrepancy and (2) limited diversity in content information. We address these issues in Sections \ref{subsec:mainstylematching} and \ref{sec:ICD}, respectively. Our proposed approach is illustrated in Figure \ref{fig:overview}.

\subsection{Style Matching}\label{subsec:mainstylematching}
Experiments in Figure \ref{fig:styleStat}a, reveal the failure of DM \cite{zhao2023dataset} in capturing the style of the original dataset. 
Furthermore, Figure \ref{fig:styleStat}c, and d illustrate the disruptive effect of this style gap on performance; in line with recent findings in deep learning community \cite{nam2021reducing,hermann2020origins,kang2022style,zhou2021domain,zhao2022style}.
Hence, we aim to enforce the condensed data to represent the style of the original large dataset. Specifically, we introduce the Style Matching (SM) module to DM \cite{zhao2023dataset} framework, comprising two complementary sub-modules: (1) Moments Matching (MM), aligning the first and second moments of feature maps, and (2) Correlation Matching (CM), aligning the correlations among feature maps.
We detail them in the following two sections.

\begin{table*}[tb]

\renewcommand\arraystretch{1}
\centering
\small

\setlength{\tabcolsep}{1.65pt}
\setlength{\abovecaptionskip}{0.1cm}
\resizebox{1\linewidth}{!}{
\begin{tabular}{c|ccc|c|c|cc|c|c|c|c|c|c}
\toprule
\multirow{2}{*}{Dataset}           & \multirow{2}{*}{IPC} & \multirow{2}{*}{Ratio\%} &  \multirow{2}{*}{Resolution} & \multicolumn{3}{c|}{Coreset Selection}   & \multicolumn{6}{c|}{Training Set Synthesis} & \multirow{2}{*}{Whole Dataset} \\ %
                            & &                & &
                            \multicolumn{1}{c}{Random}        & \multicolumn{1}{c}{Herding \cite{rebuffi2017icarl}}       & \multicolumn{1}{c|}{Forgetting \cite{toneva2019empirical}}   &
                            \multicolumn{1}{c}{{DD}$^\dagger$\cite{wang2018dataset}}  &
                            \multicolumn{1}{c}{{DG} \cite{zhao2021DC}}             & \multicolumn{1}{c}{{DSA} \cite{zhao2021DSA}} 	&	
                            \multicolumn{1}{c}{{DM} \cite{zhao2023dataset}}         &  \multicolumn{1}{c}{{CAFE} \cite{wang2022cafe}}         &\multicolumn{1}{c|}{$\text{Ours}$}
                              \\ \midrule

\multirow{3}{*}{CIFAR10}  & 1   & 0.02 &32 & 14.4$\pm$2.0  & 21.5 $\pm$ 1.2   & \multicolumn{1}{c|}{13.5 $\pm$ 1.2 }& - &  28.3 $\pm$ 0.5 & 28.8 $\pm$ 0.7 & 26.0$\pm$0.8 & \bf{31.6 $\pm$ 0.8} & \multicolumn{1}{c|}{27.9 $\pm$ 0.7} & \multirow{3}{*}{84.8 $\pm$ 0.1} \\

& 10  & 0.2&32 & 26.0 $\pm$ 1.2  & 31.6 $\pm$ 0.7 &\multicolumn{1}{c|}{ 23.3 $\pm$ 1.0} & 36.8 $\pm$ 1.2  & 44.9 $\pm$ 0.5  & 52.1 $\pm$ 0.5 & 48.9 $\pm$ 0.6 & 50.9 $\pm$ 0.5 &    \multicolumn{1}{c|}{ \bf{53.0 $\pm$ 0.2}}&\\ 

& 50  & 1 &32& 43.4 $\pm$ 1.0  & 40.4 $\pm$ 0.6 &  \multicolumn{1}{c|}{23.3 $\pm$ 1.1} & \multicolumn{1}{c|}{-} &  53.9 $\pm$ 0.5  & 60.6 $\pm$ 0.5 & 63.0 $\pm$ 0.4 & 62.3 $\pm$ 0.4 & \multicolumn{1}{c|}{\bf{65.6 $\pm$ 0.4}}& \\ 
\midrule
                                          
\multirow{3}{*}{CIFAR100} & 1   & 0.2 &32 &  4.2 $\pm$ 0.3  & 8.3 $\pm$ 0.3 &   \multicolumn{1}{c|}{4.5 $\pm$ 0.2} & \multicolumn{1}{c|}{-} &  12.8 $\pm$ 0.3  & 13.9 $\pm$ 0.3 & 11.4 $\pm$ 0.3 & \bf{14.0 $\pm$ 0.3} &   \multicolumn{1}{c|}{13.5 $\pm$ 0.2} & \multirow{3}{*}{56.2 $\pm$ 0.3}\\ 
                              
& 10  & 2 &32 & 14.6 $\pm$ 0.5  & 17.3 $\pm$ 0.3  &  \multicolumn{1}{c|}{15.1 $\pm$ 0.3 }  & \multicolumn{1}{c|}{-}  & 25.2 $\pm$ 0.3  & 32.3 $\pm$ 0.3 & 29.7 $\pm$ 0.3  & 31.5 $\pm$ 0.2&  \multicolumn{1}{c|}{\bf{33.9 $\pm$ 0.2}}& \\

& 50  & 10&32 & 30.0 $\pm$ 0.4  & 33.7 $\pm$ 0.5  &  \multicolumn{1}{c|}{-} & \multicolumn{1}{c|}{-} &  30.6 $\pm$ 0.6  &  42.8 $\pm$ 0.4 & 43.6 $\pm$ 0.4  & 42.9 $\pm$ 0.2 & \multicolumn{1}{c|}{\bf{45.3 $\pm$ 0.3}}& \\  \midrule

\multirow{3}{*}{Tiny ImageNet} & 1   & 0.2&64 &  1.4 $\pm$ 0.1  &  2.8 $\pm$ 0.2  &   \multicolumn{1}{c|}{1.6 $\pm$ 0.1 } & \multicolumn{1}{c|}{-}  &  5.3 $\pm$ 0.1 & \bf{5.7 $\pm$ 0.1} &  3.9 $\pm$ 0.2  & \multicolumn{1}{c|}{-} & \multicolumn{1}{c|}{4.9 $\pm$0.1} & \multirow{3}{*}{37.6 $\pm$ 0.4}\\ 

& 10  & 2 &64 & 5.0 $\pm$ 0.2  &  6.3 $\pm$ 0.2  & \multicolumn{1}{c|}{5.1 $\pm$ 0.2}   &\multicolumn{1}{c|}{-}  & 12.9 $\pm$ 0.1  & 16.3 $\pm$ 0.2  & 12.9 $\pm$ 0.4  & \multicolumn{1}{c|}{-} 
 &  \multicolumn{1}{c|}{\bf{17.2 $\pm$ 0.3}}\\

& 50  & 10 &64& 15.0 $\pm$ 0.4  & 16.7 $\pm$ 0.3   & \multicolumn{1}{c|}{15.0 $\pm$ 0.3}   &\multicolumn{1}{c|}{-} &  12.7 $\pm$ 0.4  & 5.1 $\pm$ 0.2  & 25.3 $\pm$ 0.2  & \multicolumn{1}{c|}{-} &    \multicolumn{1}{c|}{ \bf{27.4$\pm$0.1}}\\ 
\midrule

\multirow{3}{*}{ImageNet-1K} & 1   & 0.2&64 &  0.5$\pm$0.2  &  -  &   - & \multicolumn{1}{c|}{-}  &  - & - &  1.3$\pm$0.2  & \multicolumn{1}{c|}{-} & \textbf{2.1$\pm$0.1}  & \multirow{3}{*}{33.8±0.3}\\ 

& 10  & 2 &64 & 2.9$\pm$0.4   &  - & -   &-  & -  & -  & 5.5$\pm$0.4  & \multicolumn{1}{c|}{-} 
 &  \textbf{7.5$\pm$1.2}\\

& 50  & 10 &64& 7.1$\pm$1.5  & -  & -   &\multicolumn{1}{c|}{-} &  - & -  & 11.4$\pm$1.2  & \multicolumn{1}{c|}{-} &    \textbf{15.6$\pm$0.8}\\ 

\bottomrule
\end{tabular}}
\caption{The performance (testing accuracy \%) comparison with state-of-the-art DC and coreset selection methods. We condense the given number of  IPCs using the training set, train a DNN on the condensed set from scratch, and evaluate the network on the original testing data. Whole Dataset: the accuracy of the model trained on the whole original training set. Ratio~(\%): the ratio of condensed images to the whole training set. {DD}$^\dagger$ uses AlexNet \cite{krizhevsky2017imagenet} for CIFAR10 dataset and all other methods use ConvNet for training and evaluation. Some entries are marked as absent due to unreported values or scalability issues of optimization-based methods.}
\label{tab:sota}
\vspace{-10pt}
\end{table*}

\vspace{-5pt}
\subsubsection{Moments Matching} \label{subsubsec:SM}

Inspired by the observation in Figure \ref{fig:styleStat}a, c, and d, here, we enforce the condensed dataset $\mathcal{S}$ to capture the style of $\mathcal{T}$ in addition to its content. To this end, we utilize the first and second moments, \ie, mean and variance, of the intermediate feature maps to explicitly enforce $\mathcal{S}$ to represent the style of the $\mathcal{T}$ \cite{zhou2021domain,li2019learning,zhou2023semi}. This is done by minimizing the mean-squared distance of these moments across original and condensed datasets, in the same way as used
in the pioneering work of AdaIN \cite{huang2017arbitrary}: 
\begin{equation}\label{Lstyle}
\resizebox{0.9\linewidth}{!}{$
\begin{aligned}
    L_{\text{MM}} &= \sum_{i=0}^{c-1} \frac{1}{2}\left( 
 \sum_{l \in \mathcal{L}} \left\Vert \boldsymbol{\mu}_{\mathcal{S}_i}^{l} - \boldsymbol{\mu}_{\mathcal{T}_i}^{l} \right\Vert^2 + \sum_{l \in \mathcal{L}} \left\Vert \boldsymbol{\sigma}_{\mathcal{S}_i}^{l} - \boldsymbol{\sigma}_{\mathcal{T}_i}^{l} \right\Vert^2 \right), \\
\boldsymbol{\mu}_{\mathcal{A}}^{l} &= \frac{1}{|\mathcal{A}|}\sum_{\mathbf{a}\in \mathcal{A}}\boldsymbol{\mu}^l(\mathbf{a}), \quad \boldsymbol{\sigma}_{\mathcal{A}}^{l} = \frac{1}{|\mathcal{A}|}\sum_{\mathbf{a}\in \mathcal{A}}\boldsymbol{\sigma}^l(\mathbf{a}); \quad \mathcal{A}\in \{\mathcal{S}_i,\mathcal{T}_i\},
\end{aligned}$}
\end{equation}
where the channel-wise mean and variance of $l$-th layer are denoted by $\boldsymbol{\mu}^l \in \mathbb{R}^{n_l}$ and $\boldsymbol{\sigma}^l \in \mathbb{R}^{n_l}$, respectively.  $n_l$ represents the number of channels in the $l$-th layer of the network $\Phi_{\theta}$.
Furthermore, the outer loop over the classes $c$ is to adapt the style matching loss function in \cite{huang2017arbitrary} to the DC framework. 
The $\boldsymbol{\mu}$ and $\boldsymbol{\sigma}$ of the feature maps at a specific layer capture the style information represented in every individual channel of that layer \cite{huang2017arbitrary,gatys2015texture}.
Matching these statistics across original and condensed data reduces the gap between the style information among them without imposing rigorous spatial constraints that can reduce cross-architecture generalization \cite{deng2024exploiting,khaki2024atom,zhang2024m3d}.
We enforce first and second moments matching across multiple layers  $\mathcal{L}$ of the DNN to ensure comprehensive style matching \cite{nam2021reducing}.

\begin{table}

\renewcommand\arraystretch{1}
\centering
\scriptsize
\setlength{\tabcolsep}{1.65pt}
\setlength{\abovecaptionskip}{0.1cm}
\resizebox{1\linewidth}{!}{
\begin{tabular}{c|ccc|ccc|ccc} 
\toprule
              & \multicolumn{3}{c|}{CIFAR10}  & \multicolumn{3}{c|}{CIFAR100}   & \multicolumn{3}{c}{TinyImageNet} \\ 
\midrule
Img/Cls       & 1        & 10       & 50       & 1        & 10       & 50        & 1        & 10       & 50         \\ 
\midrule
Resolution    & \multicolumn{3}{c|}{32 $\times$ 32} & \multicolumn{3}{c|}{32 $\times$ 32} & \multicolumn{3}{c}{64 $\times$ 64} \\ 
\midrule
Random           & 10.3{$\pm$0.8} & 25.7$\pm$0.5 & 36.8$\pm$1.2 & 2.5$\pm$0.5 & 9.5$\pm$0.9 & 21.2$\pm$0.8 & 0.5$\pm$0.6 & 4.2$\pm$0.5 & 6.5$\pm$0.8 \\ \midrule
DM   \cite{zhao2023dataset}         & 19.1{$\pm$1.9} & 32.6$\pm$0.9 & 44.9$\pm$0.7 & 4.1$\pm$0.2 & 13.5$\pm$0.4 & 28.3$\pm$0.2 & 1.6$\pm$0.2 & 6.1$\pm$0.2 & 11.5$\pm$0.9 \\
$\text{Ours}$         & \textbf{22.3$\pm$0.7} & \textbf{40.9$\pm$0.6} & \textbf{51.6$\pm$0.5} & \textbf{6.3$\pm$0.3} & \textbf{21.4$\pm$0.4} & \textbf{34.0$\pm$0.2} & \textbf{2.0$\pm$0.2} & \textbf{8.6$\pm$0.4} & \textbf{15.1$\pm$0.3} \\ \midrule
Whole Dataset & \multicolumn{3}{c|}{93.07±0.1}                    & \multicolumn{3}{c|}{75.61±0.3}     & \multicolumn{3}{c}{41.45±0.4}               \\
\bottomrule
\end{tabular}}
\caption{The performance (testing accuracy \%) comparison with DM \cite{zhao2023dataset} for CIFAR10, CIFAR100, and TinyImageNet datasets by employing ResNet-18 architecture for training and evaluation.}
\label{tab:res18_part1}
\vspace{-7pt}
\end{table}

\vspace{-10pt}
\subsubsection{Correlation Matching} \label{subsubsec:FCM} 
\vspace{-2pt}
Another well-established style indicator is the one introduced by Gatys \etal \cite{gatys2015texture} consisting of correlations among feature maps \cite{gatys2016image,gatys2015neural,huang2017arbitrary}. Specifically, Gatys \etal \cite{gatys2015texture} represent the style of the input image to a DNN by the correlation between $i$-th and $j$-th filters in layer $l$.
This correlation is captured by the Gram matrix $\mathbf{G}^l \in \mathbb{R}^{n_l\times n_l}$, computed as:
\begin{equation}
\mathbf{G}^l = \Phi^l (\Phi^l)^\top
\end{equation}
where $\Phi^l \in \mathbb{R}^{n_l \times (h_l \cdot w_l)}$ represents the feature maps from layer $l$, with $n_l$ being the number of filters and $h_l \cdot w_l$ being the spatial dimensions of the feature maps.
 
We optimize the mean-squared distance between the entries of $\mathbf{G}$ across the condensed and original datasets over a set of $\mathcal{L}$ layers, providing stationary and multi-scale style feature representations \cite{gatys2016image}.
Formally, the proposed Correlation Matching (CM) loss, $L_{CM}$, is formulated as:
\begin{equation}\label{Lgram}
\resizebox{0.97\linewidth}{!}{$
\begin{aligned}
    {L}_{{CM}} =\mathbb{E}_{\theta \sim \Theta}\left[ \frac{1}{4(h_lw_l)^2n_l^2} \sum_{i=0}^{c-1} \sum_{l \in \mathcal{L}} \left(\frac{1}{|\mathcal{S}_i|}\sum_{\tbx\in \mathcal{S}_i}\mathbf{G}^{l}(\tbx) - \frac{1}{|\mathcal{T}_i|}\sum_{\bx \in \mathcal{T}_i}\mathbf{G}^{l}(\bx)\right)^2\right],
\end{aligned}$}
\end{equation}
where \(\frac{1}{4(h_lw_l)^2n_l^2}\) is the normalization factor \cite{gatys2016image,li2017demystifying}.
By minimizing Equation \ref{Lgram},
we enforce the condensed set to capture the style statistics unique to the real datasets in each class \cite{gatys2016image,gatys2015texture,zhang2022exact}. 
It is worth noting that Equation \ref{Lstyle} captures style details within each feature map, ignoring their correlations. Equation \ref{Lgram} accounts for the correlations among feature maps, complementing Equation \ref{Lstyle}. Therefore, to include style information represented in each feature map and correlation among feature maps, we define the style matching loss function as:
\begin{equation}
\resizebox{0.35\linewidth}{!}{$
\begin{aligned}
    L_S = \alpha L_{MM} +  L_{CM},
\end{aligned}$}
\end{equation}
where $\alpha$ is a balancing factor between $L_{MM}$ and $L_{CM}$. Note that $L_{MM}$ and $L_{CM}$ discard the spatial information, desired for cross-architecture generalization \cite{deng2024exploiting,khaki2024atom}.

\subsection{Intra-Class Diversity} \label{sec:ICD}
The MMD objective in DM, Equation \ref{originalDM}, supports content matching between $\mathcal{T}$ and $\mathcal{S}$ \cite{zhao2023dataset}; however, the resulting $\mathcal{S}$ suffers from limited intra-class diversity, as shown in Figure \ref{fig:styleStat}b. 
Specifically, synthesized $\mathcal{S}$ contains similar samples within each class, \ie, samples forming local clusters in the embedding space. 
It has been shown that the generalization error is bounded by the dataset diversity \cite{khaki2024atom,sener2017active,sammut2011encyclopedia}.
In other words, the more diverse the instances within a dataset, the more generalizable the model trained on that dataset will be.


\begin{algorithm}[t]
\scriptsize
\caption{Decomposed Distribution Matching in Dataset Condensation}
\label{alg:gas}
\begin{algorithmic}[1]
\Statex \textbf{Input:}  $\mathcal{T}$: Real dataset, $\Phi_{\theta}$: DNN , $\Theta$: distribution for initializing $\theta$, $\lambda \geq0$, $\alpha \geq0$, $\beta \geq0$, $t$: total training iterations
\Statex \textbf{Output:} Condensed dataset $\mathcal{S}$
\State Initialize $\mathcal{S}$ with real images from $\mathcal{T}$ 
\For{$iter=0\dots t-1 $}
\State Initialize $\Phi_{\theta}$ with $\theta \sim \Theta$;
\State Sample $\mathcal{S}_i$ from $\mathcal{S}$ $\forall i\in \{0,\dots,c-1\}$
\State Sample $\mathcal{T}_i$ from $\mathcal{T}$ $\forall i\in \{0,\dots,c-1\}$
\State Compute $L_{S}$=$\alpha L_{MM}$+$L_{CM}$
\State Compute $L_{C}$=$\beta L_{ICD}$+$L_{MMD}$ 
\State Update the synthetic dataset $\mathcal{S}\leftarrow\mathcal{S} - \eta {\nabla_{\mathcal{S}} (\lambda L_{S} + L_{C})}$
\EndFor

\end{algorithmic}
\end{algorithm}

To promote intra-class diversity, we design $L_{ICD}$ as the  Kullback–Leibler divergence among latent features of samples in $\mathcal{S}_i$. To effectively penalize samples from forming local clusters, \ie, representing similar information, and to preserve the
correct class semantics while introducing new information beneficial for model training, we enforce k-nearest neighbors constraint on the diversity criterion. Therefore, the proposed diversity criterion is as follows:
\begin{equation}\label{kll}
\resizebox{0.68\linewidth}{!}{$
\begin{aligned}
{L}_{{ICD}} = &\mathbb{E}_{\theta \sim \Theta}\left[-\sum_{i=0}^{c-1}\sum_{\tbx \in \mathcal{S}_i}{KL}(S(\Phi_{\theta}(\tbx) \| S(\overline{\mathbf{m}}_{\tbx}))\right], \\ & \text{s.t.} \: \: \overline{\mathbf{m}}_{\tbx}=\frac{1}{k}\sum_{\tbx\in \mathcal{A}_{i}^{k}}{\Phi(\tbx)},
\end{aligned}$}
\end{equation}
where $KL(a||b)$ denotes the Kullback–Leibler divergence between distributions a, and b, and $S(.)$ is the Softmax function that transforms feature vectors into probability vectors, enabling the measurement of KL divergence between features \cite{xie2016unsupervised}. $\overline{\mathbf{m}}_{\tbx}$ represents the mean feature over the set $A_i^k$, and  $A_i^k$ denotes $k$ closest intra-class synthetic instances to $\tbx$: 
\begin{equation}
\resizebox{0.60\linewidth}{!}{$
\begin{aligned}
\mathcal{A}_{i}^{k} = \{\tbx_j; \underset{\tbx_j\in \mathcal{S}_{i}, \tbx_j \neq \tbx}{\mathrm{argmin}_k} \; ||\Phi(\tbx_j) - \Phi(\tbx)||^2 \}.
\end{aligned}$}
\end{equation}
This optimization penalizes synthetic samples that cluster in the embedding space of $\Phi$, resulting in more diverse intra-class instances and efficacy in capturing the distribution of original data. 

Equation \ref{originalDM} focuses on the content matching between original and condensed data \cite{zhao2023dataset}, but ignores the diversity among instances of $\mathcal{S}$. Thus, we propose to regularize Equation \ref{originalDM} with Equation \ref{kll} as the content matching loss:
\begin{equation} 
\resizebox{0.35\linewidth}{!}{$
\begin{aligned}
L_C=\beta L_{ICD} + L_{MMD}.
\end{aligned}$}
\end{equation}
Finally, we learn the synthetic dataset by solving the following optimization problem:
\begin{equation} \label{loss_landa}
\resizebox{0.38\linewidth}{!}{$
\begin{aligned}
\mathcal{S}^{*} = \underset{\mathcal{S}}{\arg\min}\:\big(\lambda {L}_{S} + {L}_{C}),
\end{aligned}$}
\end{equation}
where $\lambda$ balances the contributions of the style matching loss $L_S$ in the overall optimization. A summary of the learning algorithm is provided in Algorithm \ref{alg:gas}.
\begin{table}

\centering
\scriptsize
\begin{tabular}{c|cc|cc} 
\toprule
               & \multicolumn{2}{c|}{ImageWoof}   & \multicolumn{2}{c}{ImageNette} \\ 
\midrule
Img/Cls          & 1        & 10 & 1        & 10               \\ 
\midrule
Resolution     & \multicolumn{2}{c|}{128 $\times$ 128}   & \multicolumn{2}{c}{128 $\times$ 128} \\ 
\midrule
Random             & 13.9$\pm$1.1 & 26.9$\pm$1.8 & 23.1$\pm$1.5 & 47.5$\pm$2.2 \\ \midrule
DM  \cite{zhao2023dataset}           & 20.9$\pm$1.5 & 31.2$\pm$0.6 & 32.5$\pm$0.4 & 55.6$\pm$0.7  \\
$\text{Ours}$        & \textbf{23.8$\pm$0.5} & \textbf{34.5$\pm$0.3} & \textbf{36.0$\pm$0.6} & \textbf{58.1$\pm$0.2}
 \\ \midrule
Whole Dataset                     & \multicolumn{2}{c|}{67.0±1.3}     & \multicolumn{2}{c}{87.4±0.1}               \\
\bottomrule
\end{tabular}
\caption{The performance (testing accuracy \%) comparison with DM \cite{zhao2023dataset} for ImageWoof and ImageNette subsets of mageNet-1K, by employing ConvNet architecture for training and evaluation.}
\label{tab:res18_part2}
\vspace{-5pt}
\end{table}

\section{Experiments}
\subsection{Datasets} \label{sec:data}

We conduct evaluation on CIFAR10, CIFAR100 \cite{krizhevsky2009learning} (32 $\times$ 32 pixels), and TinyImageNet along with ImageNet-1K \cite{deng2009imagenet} (resized to 64 $\times$ 64 pixels). 
Also, we evaluate our method on high-resolution (128$\times$128 pixels) subsets of ImageNet-1K, \ie, ImageNette and ImageWoof, containing instances from 10 classes \cite{cazenavette2022dataset}.

\subsection{Implementation details} \label{sec:imp}
We evaluate our method on $\text{IPC}\in\{1,10,50\}$
using ConvNet and ResNet-18 \cite{he2016deep}. Experimental settings and DNN architectures are consistent with DM \cite{zhao2023dataset} unless specified. 
To handle input size of 64$\times$64 and 128$\times$128 pixels, we extend the ConvNet, which has three blocks, by adding a fourth and fifth convolutional block, respectively.
We initialize $\mathcal{S}$ with randomly selected images from $\mathcal{T}$ and optimize using SGD optimizer with a fixed learning rate of 1.0. 
The differentiable augmentation strategy \cite{zhao2021DSA} is employed, as used in DM \cite{zhao2023dataset}. 
We train 20 DNNs from scratch on condensed sets with different initialization seeds, evaluating each on real test data. This process is repeated five times, resulting in five condensed datasets and 100 trained DNNs per IPC. We report the mean and variance of accuracy across these networks. The same DNN architecture is used for both training and evaluation unless specified.
Hyperparameters $\alpha$, $\beta$ and $\lambda$
are set to $1.0$, $10.0$ and $5 \times 10^{3}$, respectively, determined empirically. Also, the number of nearest neighbors for Equation \ref{kll} is set to $0.2 * \text{IPC}$. Section B of Supplementary Material provides detailed ablation on hyperparameters.

\subsection{Comparisons with State-of-the-art Methods} \label{sec:comp}
Here we compare our approach with DC baselines of DM \cite{zhao2023dataset}, CAFE~\cite{wang2022cafe}, DD~\cite{wang2018dataset}, DG~\cite{zhao2021DC}, and DSA~\cite{zhao2021DSA}, as well as coreset selection methods of Random \cite{rebuffi2017icarl}, Herding \cite{rebuffi2017icarl, castro2018end}, and Forgetting \cite{toneva2019empirical}, as shown in Table \ref{tab:sota}.
Comparing the results of DC approaches with coreset selection methods highlights the superiority of DC over coreset selection. 
Our method consistently outperforms DM across datasets and IPCs.
Particularly, with IPC=10, proposed approach surpasses DM \cite{zhao2023dataset} on CIFAR10, CIFAR100, TinyImageNet and ImageNet-1K, with considerable margins of 4.1\%, 4.2\%, 4.3\%, and 2.0\%, respectively. Concretely, with IPC=50, our method outperforms DM by 2.6\%, 1.7\%, 2.1\%, and 4.2\% in CIFAR10, CIFAR100, TinyImageNet and ImageNet-1K, respectively. 
These consistent improvements across datasets of varying sizes and resolutions underlines that the proposed method is not confined to the dataset size and resolution. 

\begin{figure}
    \centering
    \includegraphics[width=1\linewidth]{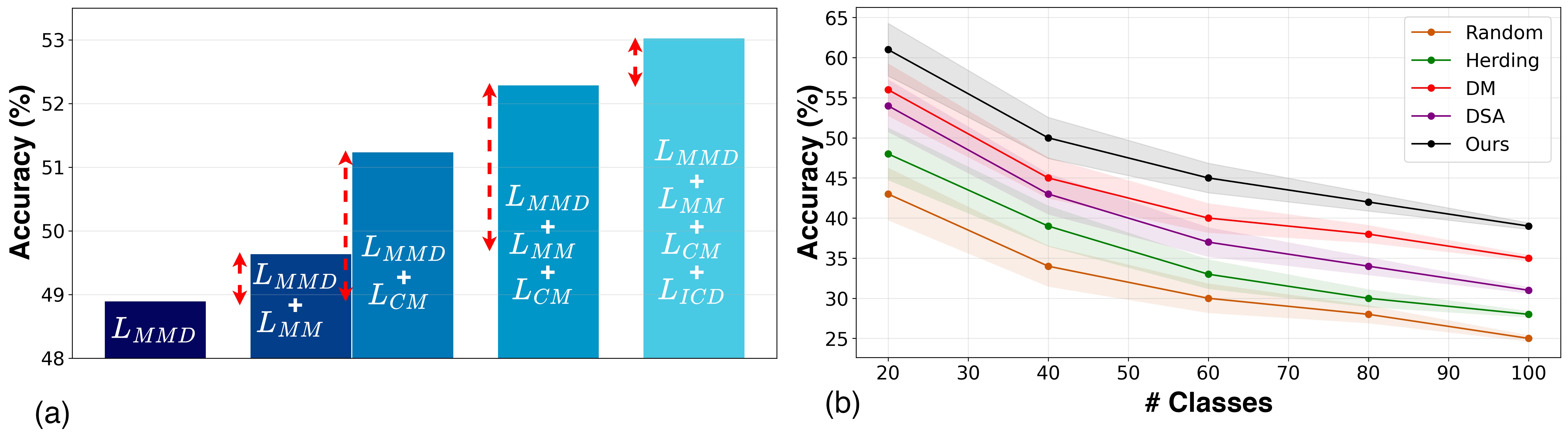}
    \caption{(a) Ablation on loss components on CIFAR10 with IPC=10 by employing ConvNet.
b) Evaluation in continual learning for CIFAR100 in five steps, \ie, 20 classs per step.
Shaded regions show the performance tolerance.
}
    \label{fig:cont}
    \vspace{-12pt}
\end{figure}

Moreover, Table \ref{tab:sota} demonstrates the consistent superiority of the proposed method over DM at $\text{IPC}=1$.
As intra-class diversity is not applicable to IPC=1, these improvements underscore the effectiveness of the proposed SM module and highlight the role of style alignment between $\mathcal{T}$ and $\mathcal{S}$.
With $\text{IPC}=10$ our method improves CAFE \cite{wang2022cafe} on CIFAR10, and CIFAR100, with noticeable margin of 2.1\%, and 2.9\%, respectively. Concretely, with $\text{IPC}=50$ our proposal outperforms CAFE on CIFAR10, and CIFAR100 with considerable margin of 3.4\%, and 2.6\%, respectively.  Please note that CAFE \cite{wang2022cafe} does not report performance on large-scale datasets such as ImageNet-1K due to its heavy computation load. These improvements across datasets and IPCs showcase the importance of style alignment and intra-class diversity which results to outperforming CAFE with fewere computational burden.

The proposed method surpasses CAFE \cite{wang2022cafe} in all IPCs except IPC=1. Concretely, improvements of our approach over DM are more pronounced at $\text{IPC}>1 $. These results highlight the importance and effectiveness of $L_{ICD}$, which is relevant only for $\text{IPC}>1$.
Table \ref{tab:res18_part1}
showcases the efficacy of our method on ResNet-18, as a more sophisticated architecture than ConvNet. Particularly, our approach outperforms DM with IPC=10 by considerable margins of 8.3\%, 7.9\%, 2.5\% in CIFAR10, CIFAR100 and TinyImageNet, respectively. In IPC=50, our method improves DM by 6.6\%, 5.7\%, and 3.6\% in CIFAR10, CIFAR100 and TinyImageNet, respectively. These improvements highlights that our method is not limited to simple architectures like ConvNet.
Again, improvements are more pronounced at $\text{IPC} > 1$, emphasizing the importance of $L_{ICD}$.
Comparing Tables \ref{tab:sota} and \ref{tab:res18_part1}, changing ConvNet to ResNet-18, \ie, more sophisticated architecture, results in more considerable improvements over DM. This suggests better generalization and practicality of the proposed method since it scales well with increased network complexity.

\begin{table}

\centering
\scriptsize
\resizebox{1.0\linewidth}{!}{
\begin{tabular}{c|c|c|c|c|c}
\toprule
\multicolumn{1}{c}{} &\multicolumn{1}{c}{} & \multicolumn{4}{c}{Test Model} \\
\cmidrule{3-6}
Method & Train Model & ConvNet & AlexNet & VGG-11 & ResNet-18 \\
\midrule
\multicolumn{1}{c|}{DSA\cite{zhao2021DSA}} &\multicolumn{1}{c|}{ConvNet} & 51.9±0.4 & 34.3±1.6 & 42.3±0.91 & 41.0±0.4 \\
\multicolumn{1}{c|}{DM\cite{zhao2023dataset}} &\multicolumn{1}{c|}{ConvNet} & 48.6±0.63 & 38.3±1.2 & 40.8±0.4 & 39.2±1.2 \\
\multicolumn{1}{c|}{CAFE\cite{wang2022cafe}} &\multicolumn{1}{c|}{ConvNet} & 50.9 ± 0.5 & 41.1±0.8 & 41.9±0.1 & 40.1±0.2 \\ 
\text{Ours}& ConvNet & \textbf{53.0±0.3} & \textbf{48.7±0.8} & \textbf{46.2±0.8} & \textbf{42.6±0.8} \\ \midrule
 \multirow{3}{*}{{$\text{Ours}$}}& AlexNet & 36.4±0.9 & 32.8±1.34 & 32.5±1.0 & 33.9±0.9 \\
 & VGG-11 & 41.2±0.4 & 37.4±0.3 & 41.7±0.4 & 38.8±0.8 \\
 & ResNet-18 & 41.9±0.5 & 34.7±1.9 & 36.65±1.0 & 40.93±0.6 \\
\bottomrule
\end{tabular}}
\caption{Cross-architecture (testing accuracy \%) performance of our proposed method compared to DM \cite{zhao2023dataset}, DSA \cite{zhao2021DSA} and CAFE \cite{wang2022cafe} methods for CIFAR10 with IPC=10.}
\label{tab:cross}
\vspace{-16pt}
\end{table}

 Table \ref{tab:res18_part2} compares our method with DM on datasets with higher resolution, \ie, $128\times128$ pixels, using ConvNet as the backbone. 
Concretely, our method outperforms DM across datasets and IPCs, showcasing that it is not restricted to low- and medium-resolution datasets. Specifically, the proposed approach improves upon DM by at least 2.9\%, and 2.5\% in ImageWoof and ImageNette, respectively. 
Consistent improvements in Tables \ref{tab:sota}, \ref{tab:res18_part1}, and \ref{tab:res18_part2} showcase that the proposed method is not confined to a specific resolution, dataset scale, or DNN architecture.

\vspace{-2pt}
\subsection{Cross-architecture Evaluations} 
\label{sec:cross}

Here we assess the cross-architecture transferability of our method by learning a condensed dataset with one architecture and evaluating it on different architectures.
To this end, we used ConvNet, AlexNet \cite{krizhevsky2012imagenet}, VGG-11 \cite{simonyan2014very}, and ResNet-18 \cite{he2016deep} architectures, as shown in Table \ref{tab:cross}. 
Using ConvNet for condensing dataset, our approach consistently outperforms its competitors across all evaluation architectures, underscoring its transferability across diverse DNN architectures. Specifically, our method outperforms DM by 10.3\%, 5.4\% and 3.4\% when testing with AlexNet, VGG-11 and ResNet-18, respectively. Also, the performance of employing ConvNet for learning condensed set surpasses more complex architectures. This result is in line with the observation of DM \cite{zhao2023dataset} that more complex architecture results in convergence issues and noisy features.

\begin{figure*}
\centering
\includegraphics[width=\textwidth]{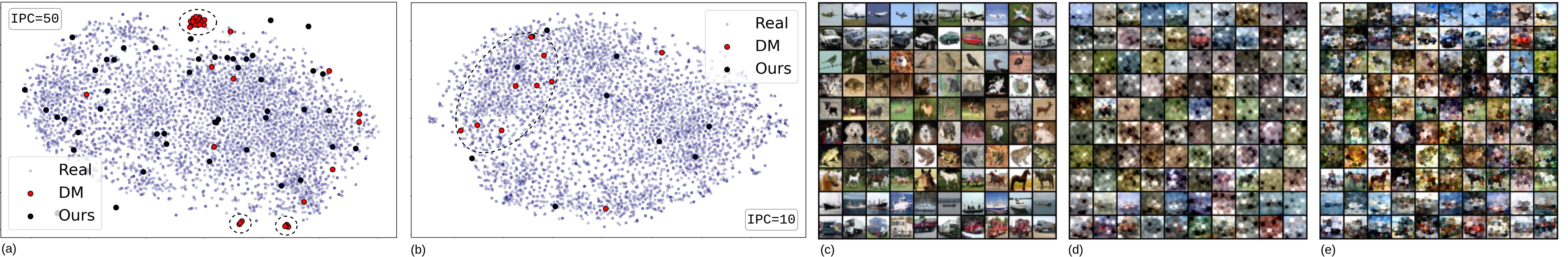}
\vspace{-20pt}
\caption{ (a,b) Intra-class diversity of two randomly selected classes of CIFAR10 with (a) IPC=50 and (b) IPC=10. Our method enhances diversity across both IPCs, addressing the limited intra-class diversity issue in DM. (c, d, e) Visualizations samples from (c) original and (d) condensed DM \cite{zhao2023dataset} and (e) our method for CIFAR10 with IPC=10. Both methods are initialized from real samples. The proposed method improves visual quality and diversity.}
\label{fig:syn}
\vspace{-10pt}
\end{figure*}

\subsection{Orthogonality to DM-based Methods}
Our proposal aims to improve distribution matching in DC by decomposing the dataset distribution into style, and content. Previous studies based on DM \cite{zhao2023dataset} overlook the significance of style alignment between original and condensed data \cite{sajedi2023datadam,zhao2023improved}. Here, we evaluate the effectiveness of our proposed SM module on two DM-based methods, as shown in Table \ref{tab:Orth}. Specifically, we modified the training code of DataDM \cite{sajedi2023datadam} and IDM \cite{zhao2023improved} to include our style alignment loss function, resulting in improved performance. 
The results indicate that our method is orthogonal to existing DM-based methods, highlighting the importance of style alignment, consistent with the well-established DNN bias toward style information \cite{nam2021reducing,hermann2020origins,kang2022style,zhou2021domain,zhao2022style}.

\begin{table}

\renewcommand\arraystretch{1}
\centering

\setlength{\tabcolsep}{1.65pt}
\setlength{\abovecaptionskip}{0.1cm}
\resizebox{1\linewidth}{!}{
\begin{tabular}{c|c|c|c|c|c|c|c|c|c} 
\toprule
              & \multicolumn{3}{c|}{CIFAR10}  & \multicolumn{3}{c|}{CIFAR100}   & \multicolumn{3}{c}{TinyImageNet} \\ 
\midrule
Img/Cls       & 1        & 10       & 50       & 1        & 10       & 50        & 1        & 10       & 50         \\ 
\midrule
Resolution    & \multicolumn{3}{c|}{32 $\times$ 32} & \multicolumn{3}{c|}{32 $\times$ 32} & \multicolumn{3}{c}{64 $\times$ 64} \\ 
\midrule
DM \cite{zhao2023dataset}            & 26.0{$\pm$0.8} & 48.9$\pm$0.6 & 63.0$\pm$0.4 & 11.4$\pm$0.3 & 29.7$\pm$0.3 & 43.6$\pm$0.4 & 3.9$\pm$0.2 & 12.9$\pm$0.4 & 25.3$\pm$0.2 \\
$\text{DM \cite{zhao2023dataset}+SM}$         & \textbf{27.8$\pm$0.7} & \textbf{52.3$\pm$0.4} & \textbf{64.1$\pm$0.7} & \textbf{13.5$\pm$0.2} & \textbf{33.0$\pm$0.1} & \textbf{45.1$\pm$0.3} & \textbf{4.91$\pm$0.1} & \textbf{16.1$\pm$0.2} & \textbf{25.2$\pm$0.3} \\ \midrule
$\text{DataDAM \cite{sajedi2023datadam}}$         & 32.0$\pm$1.2 & 54.2$\pm$0.8 & 67.0$\pm$0.4  & 14.5$\pm$0.5 & 34.8$\pm$0.5 & 49.4$\pm$0.3& 8.3$\pm$0.4 & 18.7$\pm$0.3 & 28.7$\pm$0.3 \\ 
$\text{DataDAM \cite{sajedi2023datadam}+SM}$         & \textbf{33.2$\pm$0.9} & \textbf{56.4$\pm$0.6} & \textbf{68.5$\pm$0.4} & \textbf{15.6$\pm$0.7} & \textbf{35.8$\pm$0.6} & \textbf{50.2$\pm$0.2} & \textbf{9.6$\pm$0.5} & \textbf{20.1$\pm$0.4} & \textbf{29.8$\pm$0.2} \\ \midrule 
$\text{IDM \cite{zhao2023improved}}$         &45.6$\pm$0.7&58.6$\pm$0.1& 67.5$\pm$0.1 & 20.1$\pm$0.3 & 45.1$\pm$0.1 & 50.0$\pm$0.2 & 10.1$\pm$0.2 & 21.9$\pm$0.2 & 27.7$\pm$0.3 \\
$\text{IDM \cite{zhao2023improved}+SM}$         & \textbf{46.8$\pm$0.4} & \textbf{60.2$\pm$0.2} & \textbf{68.8$\pm$0.3} & \textbf{21.6$\pm$0.4} & \textbf{47.2$\pm$0.3} & \textbf{51.3$\pm$0.4} & \textbf{11.6$\pm$0.4} & \textbf{23.8$\pm$0.5} & \textbf{28.9$\pm$0.2} \\ \midrule
Whole Dataset & \multicolumn{3}{c|}{84.8±0.1}                    & \multicolumn{3}{c|}{56.2±0.3}     & \multicolumn{3}{c}{37.6±0.4}               \\
\bottomrule
\end{tabular}}
\caption{Performance (testing accuracy \%) comparison after integrating our proposed Style Matching (SM) loss with baseline DM, and two recent DM-based methods, DataDam \cite{sajedi2023datadam} and IDM \cite{zhao2023improved}. Results are for CIFAR-10, CIFAR-100, and TinyImageNet datasets using the ConvNet architecture.}
\label{tab:Orth}
\vspace{-10pt}
\end{table}

\vspace{-2pt}
\subsection{Ablation on Loss Components}
Here, we assess the contribution of each loss component to the overall performance of our method on CIFAR10 with IPC=10. 
Results in Figure \ref{fig:cont}a reveal that both style matching supervisions, $L_{MM}$, and $L_{CM}$, improve upon the baseline $L_{MMD}$, highlighting the importance of style matching between original and condensed datasets.
Comparing (DM+MM) and (DM+CM) against (DM+MM+CM) validates the complementary nature of style information captured by mean and variance of feature maps ($L_{MM}$) and the correlation among feature maps ($L_{CM}$). 
Furthermore, incorporation of $L_{ICD}$ leads to an additional improvement, emphasizing the significance of intra-class diversity to effectively capture the real dataset distribution. 

Furthermore, Figure \ref{fig:syn}a, and b show the t-SNE visualizations of the feature distribution for two categories with IPC=10 and IPC=50. $L_{ICD}$ effectively addresses limited diversity of DM.
This advantage is consistent across IPCs, demonstrating the generalizability of the proposed $L_{ICD}$.
In addition, Figure \ref{fig:syn}c, d, and e display 10 samples per class from the real CIFAR10 dataset, and the condensed sets learned by DM and our method.
Our method improves visual quality and diversity relative to DM, highlighting the efficacy of the SM module (Section \ref{subsubsec:FCM}) and ICD (Section \ref{sec:ICD}) components, respectively, in reducing the style gap and improving the intra-class diversity. 
We provided additional visualizations for CIFAR100 and TinyImageNet in Section C of Supplementary Materials.
Also, please refer to Section D and E for ablation on style and the impact of the SM module across different blocks of ConvNet, respectively.

\vspace{-5pt}
\section{Applications: Continual Learning}
One primary motivation of DC is to mitigate catastrophic forgetting in Continual Learning (CL) \cite{rebuffi2017icarl}, making CL a reliable metric for evaluating condensation methods. To evaluate our proposal on CL, we store samples from a data stream within a predefined memory budget in a class-balanced manner.
After each memory update, the model is retrained from scratch using the latest memory, which is replaced by the condensed set while adhering to memory budget and class balance constraints. Figure \ref{fig:cont}b shows our results against the Random \cite{rebuffi2017icarl}, Herding \cite{rebuffi2017icarl, castro2018end}, DSA \cite{zhao2021DSA}, and DM.
To ensure reliability and omit the effect of class order, these experiments are repeated five times with different class orders. 
Our method outperforms its competitors 
with final test accuracy of 39.9\%, compared to 24.8\%, 28.1\%, 31.7\%, and 34.4\% for Random, Herding, DSA and DM, respectively. As the number of classes increases, the performance gap between the proposed method and the DM baseline is more evident emphasising the scalability of our method into large-scale setup.

\vspace{-5pt}
\section{Conclusion}
In this paper, we decomposed the distribution matching in DC into style and content matching. Specifically, we alleviate two shortcomings of (1) style discrepancy between original and condensed datasets, and (2) limited intra-class diversity in the condensed set, in current DC methods based on distribution matching. 
Our proposed style matching module reduces style disparity between real and condensed datasets by utilizing the first and second moments of DNN feature maps. We introduce a criterion based on KL-divergence to promote intra-class variability within the condensed dataset.
The efficacy of the proposed method is demonstrated through extensive experiments on datasets of varying sizes and resolutions, across diverse architectures, and in the application of continual learning.

{\small
\bibliographystyle{ieee_fullname}
\bibliography{egbib}
}

\end{document}


\title{Supplementary Materials of\\
Decomposed Distribution Matching in Dataset Condensation}


\author{Sahar Rahimi Malakshan, Mohammad Saeed Ebrahimi Saadabadi,\\
Ali Dabouei, and Nasser M. Nasrabadi\\
{\tt\small {sr00033, me00018, Ad0046}@mix.wvu.edu, nasser.nasrabadi@mail.wvu.edu}
}
\maketitle

\begin{figure*}
    \centering
    \includegraphics[width=1.0\linewidth]{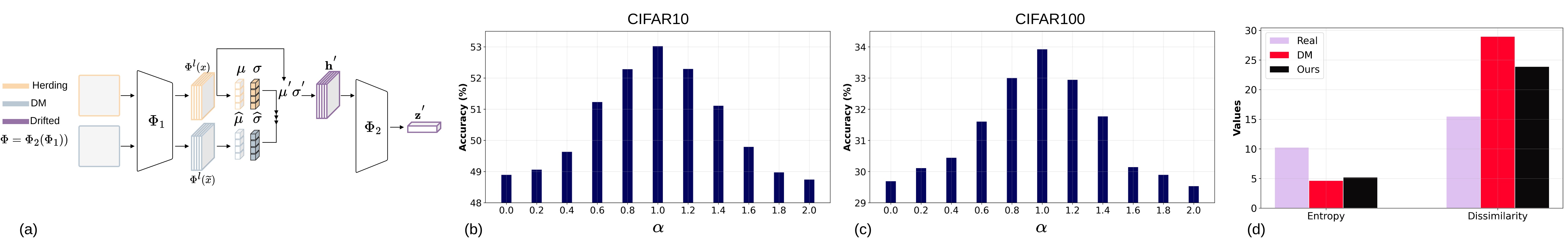}
    \caption{a) Details of the experiment in Figure \textcolor{red}{1}c, and d of the manuscript. (b, c) Ablation on $\alpha$ in Equation \textcolor{red}{7} for IPC=10 on both CIFAR10 and CIFAR100 datasets. d) Average dissimilarity and entropy texture features based on GLCM method \cite{haralick1973textural} across real and condensed set with IPC=10 for one category in CIFAR10 datasets. The texture features of the condensed set learned by our method more closely resemble those of real images, compared to the DM method.
}
    \label{fig:sup1}
\end{figure*}

\section{Impact of Style Discrepancy on DC}

To illustrate the effect of style discrepancy between the condensed and original datasets, we conduct experiments in which we drift the style of samples from Herding [\textcolor{green}{45}] coreset selection $(\boldsymbol{\mu}^l,\boldsymbol{\sigma}^l)$ 
 toward that of DM $(\widehat{\boldsymbol{\mu}}^l,\widehat{\boldsymbol{\sigma}}^l)$, as shown in Figure \textcolor{red}{1}.c, and d of the manuscript.
 Specifically, during the training of a CNN, the drifted style information is computed by a convex combination of $(\boldsymbol{\mu}^l,\boldsymbol{\sigma}^l)$ and $(\widehat{\boldsymbol{\mu}}^l,\widehat{\boldsymbol{\sigma}}^l)$:
\begin{equation}\label{sigma_new}
 \small
 \begin{aligned}
  {\boldsymbol{\sigma}}^l_{drifted} = (1-\gamma) \boldsymbol{\sigma}^l + \gamma  \widehat{\boldsymbol{\sigma}}^l,
\end{aligned}
\end{equation}
\begin{equation}\label{mu_new}
 \small
 \begin{aligned}
  {\boldsymbol{\mu}^l}_{drifted} = (1-\gamma) \boldsymbol{\mu}^l +  \gamma \widehat{\boldsymbol{\mu}}^l,
\end{aligned}
\end{equation}  
where $\gamma$ denotes the drift ratio, \ie, the extent to which the style information shifts from the original towards the target style. 
Then, we compute the feature maps with the drifted style information, following the approach of the pioneering work [\textcolor{green}{56}]:
\begin{equation}\label{adain}
 \small
 \begin{aligned}
  {\mathbf{\Phi}}^{l}_{drifted} = \sqrt{{\boldsymbol{\sigma}^l}_{drifted}} \frac{{\mathbf{\Phi}}^{l}-\boldsymbol{\mu}^l}{\sqrt{\boldsymbol{\sigma}^l}}+{\boldsymbol{\mu}^l}_{drifted}.
\end{aligned}
\end{equation}
Subsequently, $\boldsymbol{\Phi}^l_{drifted}$ passes through the remaining layers of $\boldsymbol{\Phi}$, as shown in Figure \ref{fig:sup1}a.

Figures \textcolor{red}{1}.c, and d of the manuscript show the effect of style discrepancy. As the style diverges from that of the original samples, \ie, increasing the gap between the training and testing data styles, the model performance decreases. This outcome is consistent with the well-established style bias in DNNs [\textcolor{green}{19}, \textcolor{green}{2}, \textcolor{green}{72}, \textcolor{green}{65}, \textcolor{green}{65}].

\begin{figure}
    \centering
    \includegraphics[width=1.0\linewidth]{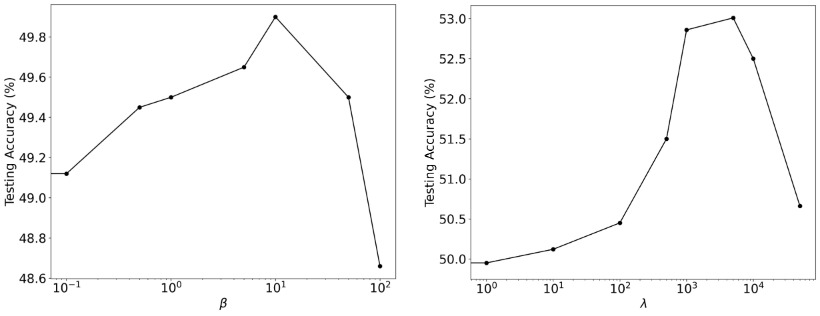}
    \caption{Ablation on $\beta$ and $\lambda$ in Equations \textcolor{red}{10} and \textcolor{red}{11} of the manuscript, respectively, for IPC=10 on CIFAR10 dataset.}
    \label{fig:betalambda}
    \vspace{-10pt}
\end{figure}

\section{Ablation on Hyperparamers}

\subsection{$\alpha$ in Equation \textcolor{red}{7}}
The overall style matching objective is defined as $L_S = \alpha L_{MM} + L_{CM}$, where $\alpha$ is a weighting factor balancing the moments matching, $L_{MM}$, and correlation matching, $L_{CM}$, losses.
Here, we perform ablation on the $\alpha$, shown in Figure \ref{fig:sup1}b, and c. 
Results show that employing both $L_{MM}$ and $L_{CM}$ with equal weight, \ie, $\alpha=1$, yields the best performance, highlighting the complementary roles of these two losses.
Specifically, $L_{MM}$ captures style information represented by the mean and variance of feature maps, while $L_{CM}$ captures style information through the correlation among feature maps.

\begin{table}
\centering
\resizebox{\columnwidth}{!}{%
    \begin{tabular}{l|c|c|c|c|c|c|c|c|c|c|c}
        \toprule
         \multicolumn{11}{c}{$k$} \\
        \cmidrule(lr){2-12}
        $IPC \times $ & $0.0$ & $0.1$ & $0.2$ & $0.3$ & $0.4$ & $0.5$ & $0.6$ & $0.7$ & $0.8$ & $0.9$ & $1$ \\
        \midrule
        IPC=10 & $48.95$ & $49.15$ & $49.90$ & $49.83$ & $49.42$ & $48.81$ & $48.14$ & $47.85$ & $47.54$ & $46.65$ & $45.20$ \\
        IPC=50 & $63.00$ & $63.56$ & $63.96$ & $63.68$ & $63.45$ & $63.14$ & $62.5$ & $61.9$ & $61.2$ & $61.15$ & $58.5$ \\
        \bottomrule
    \end{tabular}
}
\caption{Ablation study on the hyperparameter $k$ for $L_{ICD}$ in Equation \textcolor{red}{9} for IPC=10 and 50 on CIFAR10 dataset, showing the testing accuracy (\%) of the condensed dataset on CIFAR10.}
\label{tab:ablationK}
\end{table}

\subsection{$k$ in Equation \textcolor{red}{9}}

Figures \textcolor{red}{4}a, and b of the manuscript show that condensed samples learned by DM [\textcolor{green}{67}] tend to form dense clusters, indicating the need for a criterion to encourage diversity. In $L_{ICD}$, $k$ specifies the number of nearest intra-class samples in the embedding space.
We designed the loss to repel each condensed sample from its $k$ closest intra-class neighbors, thereby enhancing intra-class diversity.
We conducted experiments to determine the optimal $k$ for different IPCs. A smaller $k$ focuses on diversifying a localized neighborhood of samples, while a larger $k$ degrades results by encouraging broader dispersion. Large $k$ values can overly disperse synthetic samples, compromising class consistency and authenticity. Our experiments revealed that setting $k$ to $0.2 \times \text{IPC}$ yields optimal results for both IPC=10 and IPC=50.

\begin{figure*}

\centering
\includegraphics[width=\textwidth]{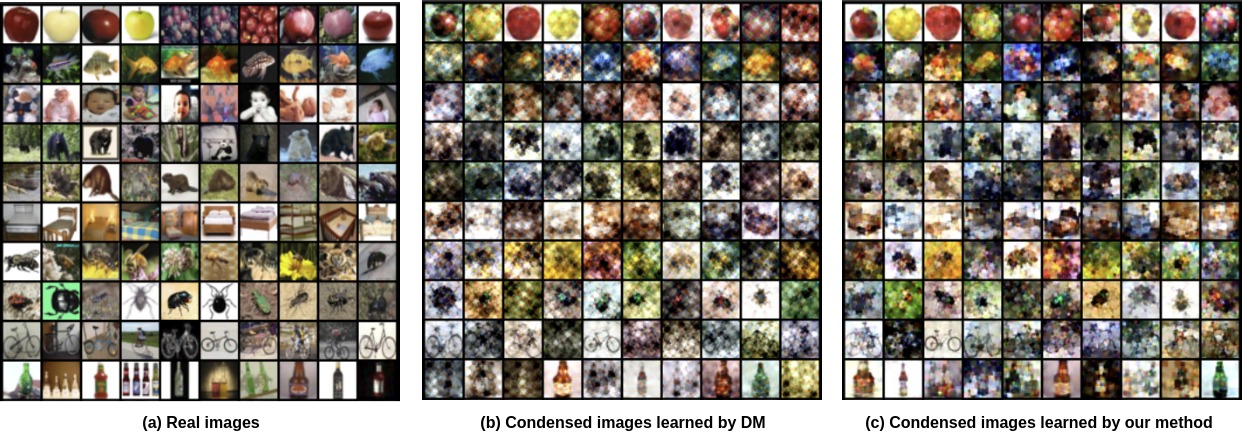}
\caption{Visualizations of (a) real and (b) condensed images learned by DM and (c) our method for CIFAR100 with IPC=10. Both methods are initialized from real samples. Our method exhibits improved visual quality and diversity compared to DM.}
\label{fig:cifar100}
\vspace{-7pt}
\end{figure*}
\subsection{$\beta$ in Equation \textcolor{red}{10} and $\lambda$ in Equation \textcolor{red}{11}}

Figure \ref{fig:betalambda} illustrates the impact of $\beta$ and $\lambda$ on our method's performance, corresponding to $L_{ICD}$ and $L_{S}$ in Equations \textcolor{red}{10} and \textcolor{red}{11}, respectively.
Optimal results for both loss components are achieved at $\beta=10$ and $\lambda=5\times 10^3$.
The magnitudes of $L_{ICD}$ and $L_{S}$ are significantly lower compared to $L_{MMD}$, necessitating the adjustment of hyperparameters to higher values for balance. Results in Figure \ref{fig:betalambda} demonstrate that integrating style information and promoting intra-class diversity consistently enhances performance, up to a threshold of $5\times10^3$ and $10$, respectively.
Beyond this point, performance starts to decline, attributed to an overemphasis on style matching at the expense of the discriminative features highlighted by $L_{MMD}$. Moreover, it is vital to balance intra-class diversity enhancement to prevent class overlap or confusion.
Therefore, exceeding the optimal thresholds for the style-matching and intra-class diversity coefficients results in a decline in model performance.

\section{Visualization}

\begin{figure*}
\centering
\includegraphics[width=\textwidth]{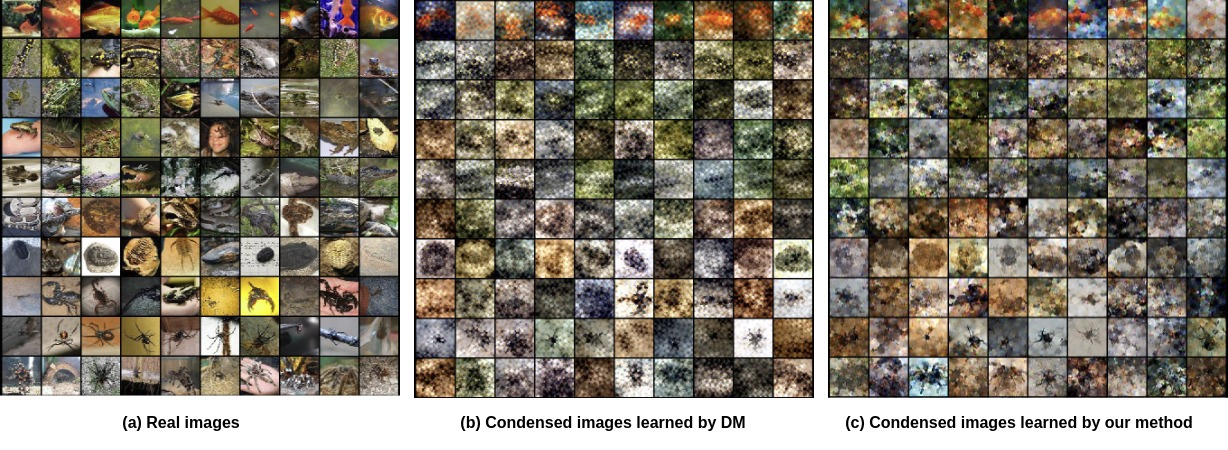}
\caption{Visualizations of (a) real and (b) condensed images learned by DM and (c) our method for TinyImageNet with IPC=10. Both methods are initialized from real samples. Our method exhibits improved visual quality and diversity compared to DM.
}
\label{fig:tiny}
\end{figure*}

Figures \ref{fig:cifar100} and \ref{fig:tiny} display the resulting condensed sets for CIFAR100 and TinyImageNet, learned by DM and our method, alongside the real images. The improvement in visual quality and diversity with our method is attributed to the SM module and ICD component, detailed in Sections 3.3 and 3.4 of the manuscript, which effectively reduce the style gap between original and condensed sets and enhance intra-class diversity among condensed samples, respectively.

\section{Style}
\subsection{Style Gap Analysis}

\begin{figure}
  \centering
  
  \includegraphics[width=1\linewidth=]{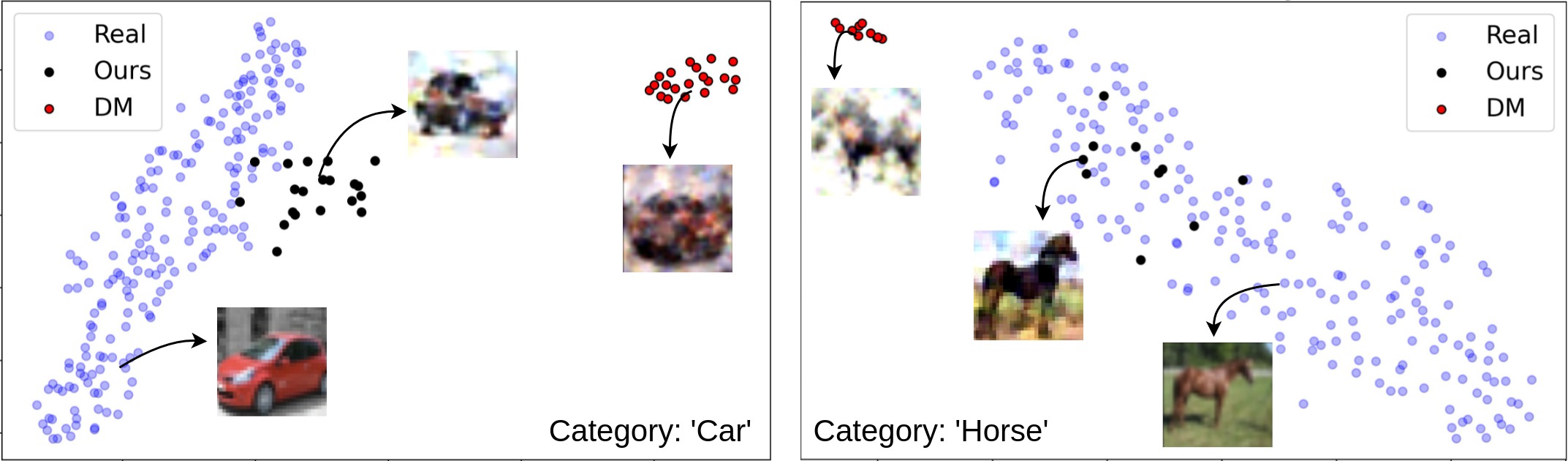}
  \caption{
2D t-SNE visualization of style statistics computed from the first layer's feature map of ConvNet,
for real CIFAR10 images, and condensed set learned by DM  and our method for two categories, demonstrating the effectiveness of our approach in reducing the style gap.}
  \label{fig:styleOurs}
  \vspace{-10pt}
\end{figure}

As discussed in the Introduction, our comparison of style indicators between CIFAR10's real and condensed datasets (Figure \textcolor{red}{1}.a) reveals a significant style gap. To evaluate our method's effectiveness in mitigating this gap, we repeated the experiment with our approach, as shown in Figure \ref{fig:styleOurs}. The results demonstrate that our method successfully narrows the style discrepancy using the SM module.

\subsection{Texture Analysis }
Conventionally, style can be characterized by the textural attributes of an image, which include roughness, smoothness, and color diversity in the image [\textcolor{green}{16}, \textcolor{green}{18}]. Texture analysis in the field of image processing is a crucial component and can be broadly categorized into four main approaches: statistical, geometric, model-based, and signal processing techniques \cite{chen1994handbook, haralick1973textural}.
Among these, the Gray-Level Co-occurrence Matrix (GLCM), introduced by Haralick \etal \cite{haralick1973textural}, is a prominent statistical method 
for texture analysis.
GLCM is foundational for texture analysis, emphasizing the spatial distribution and relation of pixels to describe an image's surface characteristics effectively \cite{haralick1973textural, szantoi2013analyzing}.



Utilizing the GLCM method, we employ two texture features including dissimilarity and entropy to analyze the textural statistics of images, which are computed as \cite{szantoi2013analyzing, kwak2019impact}:
\begin{equation}
\resizebox{0.55\linewidth}{!}{$
\begin{aligned}
    \text{Dissimilarity} &= \sum_{i=0}^{n-1} \sum_{j=0}^{n-1} p(i, j) \cdot |i - j| ,
\end{aligned}$}
\end{equation}
\begin{equation}
\resizebox{0.55\linewidth}{!}{$
\begin{aligned}
    \text{Entropy} &= -\sum_{i=0}^{n-1} \sum_{j=0}^{n-1} p(i, j) \cdot \ln(p(i, j))  ,
\end{aligned}$}
\end{equation}
where $n$ denotes the grayscale level, and $p(i, j)$ is the normalized grayscale value at positions \( i \) and \( j \) within the kernel, summing to 1. We employ different kernels ($3\times3$ and $5\times5$, a region or a set of neighbors around a central pixel) and report the average of them in the results. Dissimilarity evaluates the variation in intensity among adjacent pixel pairs, offering insights into texture contrast and complexity \cite{kwak2019impact}. Entropy, measures the randomness in intensity distribution, thereby reflecting the unpredictability and diversity of textural patterns \cite{kwak2019impact}.

As illustrated in Figure \ref{fig:sup1}d, there is a significant gap in both texture features between real images and those learned by the DM. The usage of the style matching module introduced by our method brings the texture features in the condensed set closer to real data compared to the baseline of DM [\textcolor{green}{67}], as shown in Figure \ref{fig:sup1}d. Specifically, our method achieves dissimilarity and entropy features that are 5\% and 0.56\% closer to real features compared to DM, respectively, indicating improvements in texture matching between original and learned condensed sets in our method.

\section{Style Matching in Multiple Layers }
To evaluate the impact of the SM module across different blocks, we applied it to each block of the ConvNet architecture, which consists of three convolutional blocks. Our results, presented in Table \ref{tab:layers}, indicate that applying this module individually after each block improves performance. These consistent enhancements across different blocks highlight the presence of beneficial style knowledge for DC at various depths within the DNN. Ultimately, applying this module across all three blocks yields the best results, as demonstrated in Table \ref{tab:layers}, underscoring the existence of distinct style information throughout the layers of the DNN.

\section{Application: Neural Architecture Search}

Neural Architecture Search (NAS) aims to identify the best DNN architecture candidates. NAS has become an important use case for dataset condensation (DC) since a condensed dataset can be used as a proxy for the original data to efficiently search for optimal architectures. Here, we compare the performance of the proposed method with three baselines: DM, DSA, and Random Selection. Following [\textcolor{green}{68}], we explore the application of our method in NAS on the CIFAR-10 dataset, using a search space of 720 ConvNets by varying hyperparameters. Please refer to [\textcolor{green}{68}] for full experimental details. We trained architectures on both the original and condensed datasets for 200 epochs. Table \ref{tab:NAS} presents: 1) accuracy on the test data, 2) Spearman's rank correlation coefficient between the testing accuracy of the top models selected using condensed datasets and the whole training data, 3) training time required for training 720 architectures, and 4) memory footprint of the datasets. The proposed method achieves the highest accuracy among its competitors, coming within one percent of the accuracy obtained by training on the full CIFAR-10 dataset. Moreover, the training time is significantly reduced from 8604.3 minutes to 142.6 minutes. Additionally, our method enhances the Spearman's rank correlation coefficient for DM, indicating that a reliable ranking of architectures is obtained using the proposed method.

\begin{table}[h]
  \centering
  \setlength{\abovecaptionskip}{0.1cm}
  \resizebox{0.48\textwidth}{!}{
  \begin{tabular}{l|ccc|c|c}
\toprule
                & Random & DSA   & DM    & Ours  & Whole Dataset \\ \midrule
Accuracy     & 84.0   & 82.6  & 82.8  & \textbf{84.2}  & \textbf{85.9}          \\
Correlation     & -0.04  & 0.68  & 0.76  & \textbf{0.80}  & 1.0           \\
Time cost (min) & \textbf{142.6}  & \textbf{142.6} & \textbf{142.6} & \textbf{142.6} & 3580.2        \\
Storage (imgs)  & \textbf{500}    & \textbf{500}   & \textbf{500}   & \textbf{500}   & 50000         \\ \bottomrule
\end{tabular}
  }
  \caption{Neural architecture search experiments on CIFAR-10
dataset for the search space of 720 ConvNets.}
  \vspace{-10pt}
    \label{tab:NAS}
\end{table}

\begin{table}

\footnotesize
    \centering
\begin{tabular}{cccccc|ccc|}
	\toprule
	Block 1 & Block 2 & Block 3  &  Accuracy     \\
	\midrule
 -& - &  - &  48.9 $\pm$ 0.6 \\
	\checkmark & - &  - & 50.93 $\pm$ 0.66 \\
	 -&\checkmark  & -  &  51.60 $\pm$ 0.57\\
	 -& - &  \checkmark & 51.91 $\pm$ 0.56\\
  \checkmark& \checkmark &\checkmark   & 52.29 $\pm$ 0.42\\

\bottomrule
\end{tabular}
    
    \caption{Ablation on SM module Across ConvNet convolutional blocks for CIFAR10 dataset with IPC=10.
}
\label{tab:layers}
\end{table}

{\small
\bibliographystyle{ieee_fullname}
\bibliography{egbib}
}